\documentclass[11pt]{article}

\usepackage[]{acl}

\usepackage{times}
\usepackage{latexsym}
\usepackage{graphicx} 
\usepackage{amsmath}
\usepackage{multirow}
\usepackage[T1]{fontenc}

\usepackage[utf8]{inputenc}
\usepackage{cancel}
\usepackage{soul}

\usepackage{microtype}

\usepackage[disable]{todonotes}
\usepackage{booktabs}
\usepackage{enumitem}
\usepackage{fontawesome}
\usepackage{xspace}

\newcommand\PaperTitle{Factual Confidence of LLMs: \\on Reliability and Robustness of Current Estimators}

\newcommand\pIK{$P(\text{IK})$\xspace}
\newcommand\pTrue{$P(\text{T})$\xspace}

\title{\PaperTitle}

\author{Mat\'{e}o Mahaut\thanks{{\hspace{0.1cm}}Work conducted during an internship at AWS AI Labs.}  \\
  Universitat Pompeu Fabra\\
  \texttt{mateo.mahaut@upf.edu} \\\AND
  Laura Aina \quad  Paula Czarnowska \quad  Momchil Hardalov \\
  \textbf{Thomas M\"{u}ller \quad  Llu\'{i}s M\`arquez}  \\
 AWS AI Labs \\
  \texttt{\{eailaura, czarpaul, momchilh, thomul, lluismv\}@amazon.com} \\}

\begin{document}
\maketitle
\begin{abstract}

Large Language Models (LLMs) tend to be unreliable in the factuality of their answers.
To address this problem, NLP researchers have proposed a range of techniques to estimate LLM's confidence over facts.  However, due to the lack of a systematic comparison, it is not clear how the different methods compare to one another. To fill this gap, we present a survey and empirical comparison of estimators of factual confidence. We define an experimental framework allowing for fair comparison, covering both fact-verification and question answering. Our experiments across a series of LLMs indicate that trained hidden-state probes provide the most reliable confidence estimates, albeit at the expense of requiring access to weights and training data. We also conduct a deeper assessment of factual confidence by measuring the consistency of model behavior under meaning-preserving variations in the input. We find that the  confidence of LLMs is often unstable across semantically equivalent inputs, suggesting that there is much room for improvement of the stability of models' parametric knowledge. Our code is available at \href{https://github.com/amazon-science/factual-confidence-of-llms}{https://github.com/amazon-science/factual-confidence-of-llms}.
\end{abstract}

\section{Introduction} \label{sec:intro}
A major problem of Large Language Models (LLMs) is that they do not always generate truthful information. Models can hallucinate by convincingly reporting information that is actually false or they are not confident about, or provide factual answers only when prompted in a certain way~\cite{Elazar2021MeasuringAI, lin-etal-2022-truthfulqa, 10.1145/3571730, wang2023survey,luo-etal-2023-systematic}. This behavior can be severely harmful, especially given the current explosion of LLM usage: a lack of truthfulness can lead to spread of misinformation and breaches to user trust~\cite{weidinger2021ethical,10.1145/3442188.3445922,evans2021truthful,tamkin2021understanding}. Having a reliable estimate of the model's confidence over a fact---the degree to which it is expected to have accurate factual knowledge with respect to an input---is key for mitigating this problem~\cite{geng2023survey,tonmoy2024comprehensive}.

\begin{figure}[!t]
    \centering
    \includegraphics[width=0.95\columnwidth]{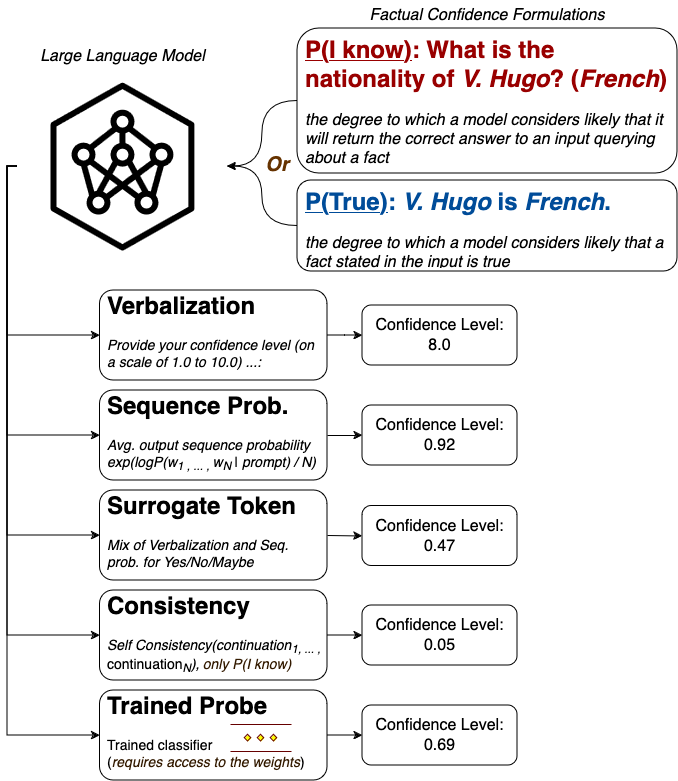}
    \caption{{Overview of our factual confidence estimation framework. We work with five groups of {methods} and two
 formulations: $P(\text{I know})$, which applies to questions, and $P(\text{True})$, which applies to statements. All of the methods produce a continuous score, except \emph{verbalization}, where the model generates a confidence level. 
 }}
    \label{fig:intro}
    \vspace{-0.5cm}
\end{figure}

Recently, a number of methodologies for estimating the factual confidence of LLMs were proposed~(\citealt{Lin2022-ni, burns2023discovering,Kuhn2023-nz, Azaria2023-ii, Pacchiardi2023-se}, among others). However, none of them establishes a unified experimental framework to compare methods. This leaves open questions regarding how aligned the methods are in their estimates, and which are the most reliable ones across models.

We aim to fill this gap by providing a survey of the current state-of-the-art for estimating the factual confidence in LLMs, and performing a systematic empirical comparison of the reliability and robustness of the existing methods. 

We introduce an %
experimental framework, shown in Figure~\ref{fig:intro}, enabling a fair comparison between methods across models and datasets. 

We adopt two distinct formulations for measuring factual confidence: 
(\emph{i})~the probability of a statement to be true, noted $P(\text{True})$---fact-verification~\citep{thorne-etal-2018-fact, Azaria2023-ii}, and (\emph{ii})~the probability of yielding a truthful answer to a query, noted $P(\text{I know})$---question answering~\citep{Kadavath2022-pf,yin2023large}. 
Additionally, we categorize the existing methods into five groups: trained probes, sequence probability, verbalization, surrogate token probability, and consistency-based.

Our experiments across eight publicly available LLMs indicate that prompting-based methods are less reliable than supervised-probing, although the latter requires training data and access to models' weights. For instruction-tuned LLMs, \emph{verbalization} and \emph{consistency-based} methods
are viable alternatives.

Further, we argue that all methods for estimating factual confidence can ultimately lead to misleading conclusions if only tested on a single way of asserting a fact: An LLM may seem to know a fact given an input, but then contradict itself given an alternative phrasing of that fact~\cite{Elazar2021MeasuringAI, kassner2021multilingual, lin-etal-2022-truthfulqa,qi2023cross, Kuhn2023-nz}. In our experiments, we find evidence of such instability, suggesting that 
LLMs do not always encode facts based on abstractions over diverse input variations.

In summary, this paper provides the following contributions:
\begin{itemize}[noitemsep]
\item A survey of the literature on LLM factual confidence estimation; 
\item An experimental framework enabling a fair comparison across methods proposed in the literature;

\item Insights about the reliability and robustness of different types of methods, providing recommendations for NLP practitioners; 

\end{itemize}

\section{Factual Confidence: Key Concepts} \label{sec:fact_conf}

\subsection{Definition of a Fact}
\label{subsec:fact}
We consider a \textit{fact} to be a piece of information that accurately represents a world state.\footnotemark A natural-language statement is \emph{truthful}---or \emph{factual}---if its meaning reports a state of affairs that is supported by a true fact: e.g.,~``\emph{Paris is a city in France}'' is truthful as the city of Paris is indeed located in France. Facts and natural-language statements are not linked by a one-to-one relation: The same fact can be declared with multiple statements, varying on the surface level, but sharing the same meaning.\footnotetext{For simplicity, in this work, we restrict our focus to minimal, atomic facts, in the sense that they do not involve a combination of other facts; e.g., ``\emph{The Louvre is in Paris}'' as opposed to ``\emph{The Louvre is in Paris, which is in France}''.}

For this reason, one's confidence in a fact should be consistent across meaning-preserving linguistic variations, such as paraphrases or translations of a statement: If we are certain that ``\emph{Paris is a city in France}'' is true, we will not doubt that its paraphrase ``\emph{Paris is a French city}'' or its translation in French (if we understand French) are also true.

\subsection{Factual Confidence} 
\label{subsec:confidence_def}
We distinguish between two {facets of factual confidence of LLMs, following~\citet{Kadavath2022-pf}}:

\vspace{0.5em}
\noindent\textbf{$P(\text{True})$} -- shortened as \pTrue: the degree to which a model considers likely that a fact stated in the input is true. For example,~``\emph{Paris is the capital of France}'' should get a high \pTrue as it is truthful and is common knowledge, while ``\emph{Sidney is the capital of France}'' should get a low \pTrue.  To estimate \pTrue scores we need to pass a statement in the input, which is evaluated in its truthfulness: this is in line with the setup of \textbf{fact-verification}~\citep{thorne-etal-2018-fact,hardalov-etal-2022-survey,guo-etal-2022-survey}.

\vspace{0.5em}
\noindent\textbf{$P(\text{I Know})$} -- shortened as \pIK: the degree to which a model considers likely that it will return the correct answer to an input querying about a fact. For instance, we can compute \pIK in a \textbf{QA} setup passing a question as input---e.g.,~``\emph{What is the capital of France}?''. If confident to know the true answer, \pIK should be high; it should instead be low in case of uncertainty. In contrast to \pTrue, \pIK is estimated without stating the fact in the input, but rather expecting a factual answer by the model complementing the query. 

\vspace{0.5em}\pTrue and \pIK are both telling of the underlying factual confidence of an LLM. However, depending on the data format (e.g.,~statements vs.~questions) or task of interest (e.g.,~fact-verification vs.~QA) focusing on one of the measures is more suitable. Previous works introducing methods to estimate factual confidence have typically addressed only one of the two measures. However, as we demonstrate with our experimental framework, most methods can be adapted to estimate both \pTrue and \pIK, although in practice they may not be equally reliable in each setup.
\begin{table*}[t]
\centering

\begin{tabular}{lllll}
\toprule
\textbf{} & \textbf{Black-box} & \textbf{Trained} & \textbf{Prompt-based} & \textbf{Scores for} \\ \midrule
\textbf{Trained Probe} & No & Yes & No & \pTrue \& \pIK \\
\textbf{Sequence Probability} & Yes (*) & No & No & \pTrue \& \pIK \\
\textbf{Verbalization} & Yes & No & Yes & \pTrue \& \pIK \\
\textbf{Surrogate Token Probability} & Yes (*) & No & Yes & \pTrue \& \pIK \\
\textbf{Consistency} & Yes & No & No & \pIK \\
\bottomrule
\end{tabular}
\caption{Differences across types of factual confidence estimators. \emph{Black-box} marks methods which do not rely on access to model's weights; (*) denotes the possibility to use sampling if token probabilities are not available. 
}
\label{tab:methods}
\end{table*}

\subsection{Robustness of Factual Knowledge}
\label{subsec:robustness}
We work from the hypothesis previously voiced by~\citet{Petroni2019-pa} that a language model's factual knowledge may stem from encoding facts in its weights (\emph{parametric memory}) as an abstraction over the linguistic input in the training data. 

Such human-like robustness and abstraction ability cannot however be taken for granted~\citep{bender-koller-2020-climbing,doi:10.1073/pnas.2215907120,mahowald2024dissociating}. Testing for consistency to meaning-preserving variations of an input is key to distinguishing whether a model has encoded a fact as an abstraction over linguistic forms, as opposed to memorizing statements asserting the fact~\citep{carlini2023quantifying}. For instance, if a model has a robust encoding in its parametric memory of what the capital of France is, it should provide the same answer to ``\emph{What is the capital of France}?'', ``\emph{What is the name of the French capital city}?'' or any other rewording. Prior works already provided evidence that models may not always act consistently across semantically equivalent inputs~\citep{Elazar2021MeasuringAI,kassner2021multilingual,ohmer2023separating, qi2023cross}. However, this has not yet been investigated in relation to the degree of factual confidence.

\section{Factual Confidence: Survey of Methods} \label{sec:survey}

Based on a review of the research area, we identify five groups of existing methods to estimate factual confidence, which we discuss in the following subsections. In Table \ref{tab:methods}, we provide an overview of the functional differences among these methods. 

\subsection{Trained Probes}
The methods in this group are based on probes which compute a transformation of a model's internal representations. The probes can take as input the final layer or earlier layers, taking advantage of the latent compression stages of an LLM~\cite{Voita2019-mu}. 

\citet{Azaria2023-ii} proposed to train multi-layered probes to extract factual confidence scores from hidden states, under the argument that such estimates are less subject to surface-level features---how a claim is phrased---and thus more reliable.  Their setup is in line with an estimate of \pTrue.

\citet{Kadavath2022-pf} adopted this method in a QA setup, estimating \pIK using a trained value head on top of the final transformer layer. Breaking off from the constraints of supervised training, \citet{burns2023discovering} propose another version of the probe, which they train in an unsupervised manner, by maximizing distance between representations of contradicting answers on a Yes/No question dataset. They report performance only slightly lower than supervised alternatives---we therefore do not separate it in its own group. 
 
This method is worth considering in the specific case where a model's layer outputs are available but reliable annotations are not available to train a probe.

We must emphasize that these methods have more strict requirements compared to the other groups, as: (\emph{i})~they require having access to the model weights, and (\emph{ii})~they need supervised training data, i.e.,~data with labels that reflect if the statements are true, for \pTrue, and data with labels that reflect whether the model will provide the correct answer to the question, for \pIK.

\subsection{Sequence Probability}
\label{subsec:seqprob}
This family of methods use the averaged log probabilities, assigned to a sequence of output tokens, to estimate factual confidence. 
{For this approach to work well, sequence probabilities need to be well calibrated~\cite{guo2017calibration} to indicate probability of correctness~\cite{10.1162/tacl_a_00330, xiong2024can}.}

In the context of factual knowledge, sequence probability has been applied both in cloze tasks and QA setups~\cite{jiang-etal-2021-know,yin2023large}, which corresponds to measuring \pIK. 

A major limitation of the sequence probabilities is that they represent the confidence over \emph{how} a claim is made (i.e.,~{the probability of the generated response as a sequence of tokens---different surface realizations of the claim would have different probabilities}), rather than the confidence about the claim itself~\citep{Lin2022-ni}. Moreover, \citet{gal2016dropout} showed that the output sequence probabilities produce unreliable, overconfident estimates in general. Thereby, these methods are mainly used as a weak baseline for model confidence estimation.

\subsection{Verbalization}
\label{subsec:verbalization}

In the verbalization (\emph{verbalized confidence}) methods~\cite{Lin2022-ni, xiong2024can, yin2023large, Tian2023-wp, Kadavath2022-pf}, the model is directly prompted to report its confidence level (e.g.,~``\emph{How confident are you that the answer is correct}?''). {Here, differently from the other methods, the LLM generates a numeric confidence level as a sequence of output tokens.} 
\citet{Lin2022-ni} and \citet{Tian2023-wp} find that this method provides well-calibrated and surprisingly accurate estimates for highly capable instruction-tuned models like GPT-3~\cite{ouyang2022training, bai2022training,openai2023gpt}.

\citet{Tian2023-wp}~extend the verbalization to ``top-k'' prompting, i.e.,~prompting the model for $k$ answers, along with their estimated confidence. They also propose to use Chain-of-Thought (CoT) and multi-step prompting, e.g.,~first providing an answer and then providing a measure of confidence that this answer is correct. In this work, we focus on the simplified non-CoT, $k=1$ prompting as it shows competitive results to the more complex methods for many model-dataset-metric combinations.

\subsection{Surrogate Token Probability} \label{subsec:surrogate}
These methods, extensively studied by~\citet{Kadavath2022-pf, xiong2024can}, can be considered a hybrid approach between the \emph{Sequence Probability} (Section~\ref{subsec:verbalization}) and \emph{Verbalization} (Section~\ref{subsec:seqprob}). The input prompt asks the model to provide as output specific tokens to report the factuality of the claim in the input; the probabilities assigned to those tokens are then used to determine the confidence level. This method can be adapted to measure both \pTrue and \pIK~\cite{Kadavath2022-pf}.

\subsection{Output Consistency} 
\label{subsec:output-consistency}
\emph{Output consistency} methods (also referred to as \emph{self-consistency})~\cite{wang2023selfconsistency} build on the assumption that the high confidence of a LLM leads to generating consistent outputs. Given a question or an incomplete statement, we sample multiple completions and take the inter-responses consistency as the confidence measure. If the model consistently generates the same answer, the confidence is high. Conversely, if the model generates contradictory answers, the confidence is low. One limitation of this method is that, due to its dependence on completion generation, it can only be used to estimate \pIK and not \pTrue. 

\citet{Manakul2023-pe} demonstrated the efficacy of this method when applied to factual knowledge, focusing on GPT models and using output consistency to validate model responses.  
\citet{Kuhn2023-nz} argued that additional clustering of the outputs that are semantically equivalent as instances of the same answer is needed.

\section{Methodology} \label{sec:methodology}

\subsection{Data} \label{subsec:data}
 
Currently, there is no standardized set of benchmarking datasets that are adopted across previous work. To fill this gap, we adapt two publicly available datasets, namely Lama T-REx~\cite{Petroni2019-pa} and PopQA~\cite{mallen-etal-2023-trust}, to test factual confidence in both fact-verification (\pTrue) and QA (\pIK) setups. We believe that establishing such a baseline setup is important both for researchers and practitioners.

\subsubsection{\texorpdfstring{\pTrue}~ in Fact Verification: Lama T-REx} \label{subsubsec:fact-veri}
Lama T-REx~\cite{Petroni2019-pa} is a relational dataset made of triplets extracted from Wikipedia <subject, relation, object>, (e.g., \emph{<Victor Hugo, was born in, France>}). We use this dataset to create both true and false statements for estimating \pTrue.

We create false versions of each factual statement, by randomly substituting the object in the triplet with one from the same relation (``Victor Hugo was born in China''). This ensures the right entity type and avoids grammatical errors. 

There are 34K triplets (true statements) in the T-REx dataset. For each true statement we sample one false, creating a balanced set of 50/50 true/false statements. Then, we take 80\% of the examples, 27K of the positives and their corresponding negatives, 54K in total, for training (only used for \emph{trained probe}). The rest, we use for analysis---6.8K T-REx true statements and an equal number of false statements, 13.6K in total. 

\subsubsection{\texorpdfstring{\pIK}~ in QA: PopQA} 
\label{subsubsec:popqa}
The PopQA dataset~\cite{mallen-etal-2023-trust} consists of short questions and single entity answers (e.g.,~question: \emph{What is George Rankin's occupation?}, answer: \emph{Politician}.). 

{This dataset offers a set of synonymous phrases for each correct answer. This is an important feature, which makes the evaluation more robust to answer phrasing, and, in turn, lowers the risk of underestimating model's correctness.}

Moreover, this dataset covers a broad range of entities, with varying degrees of popularity (estimated based on the number of Wikipedia page views). On one hand, this ensures the diversity of the target entities, and, on the other, it allows for further analysis between popularity and estimated confidence, which we leave for future work.

We use PopQA to test models' factual confidence given a fact-related query, i.e.,~\pIK. 
The dataset contains 14K questions: we keep 80\% (11K) for training, and 20\% (2.8K) for testing.
By definition (Section~\ref{subsec:confidence_def}), the gold labels for \pIK should indicate if the model outputs a correct answer. Ultimately, a model's answer depends on the decoding strategy; in this work, for simplicity and clarity of interpretation, we use greedy decoding. If the answer is correct, we set the gold \pIK to 1, else to 0 (more details in Section~\ref{subsec:scoring}).
As the labels depend on model correctness, the data will have varying proportions of positive labels across models, in our case, ranging from $\sim$11\% to $\sim$27\%.\footnote{\label{footnote:popqa_acc} 
The questions from PopQA are generally considered hard (ChatGPT: 30\% accuracy, SelfRAG~\cite{asai2024selfrag}: 55\%).}

\subsection{Scoring Methods Implementation} \label{subsec:scoring}

Below, we report the main specifics of our implementation of the methods (details in Appendix~\ref{app:implementation}). 

\subsubsection{Estimating \texorpdfstring{\pTrue}~}

Given a statement, we compute \pTrue as follows: 

\noindent\textbf{Trained probe}: As in \citet{Azaria2023-ii}, we train a 3-layer fully connected feed-forward network for 10 epochs, passing as input the hidden states from the 24th transformer layer for each LLM (requires hyper-parameter optimization, later layers before the last one work better) and predicting whether a statement is true or false. 

\noindent\textbf{Sequence Probability:} Average log-probability of the statement's tokens. 

\noindent\textbf{Verbalization}: Prompting for the confidence level that the statement is true (Appendix~\ref{app:implementation}). 

\noindent\textbf{Surrogate Token Probability}: Log-probability of the ``Yes'' token following a query on whether the statement is true.

\subsection{Estimating \texorpdfstring{\pIK}~}
Below, we describe how the \pIK estimates for each method group are computed, based exclusively on the question. 

\noindent\textbf{Trained Probe:} We use the same approach as for \pTrue, but train the probes to predict whether the model's greedy-generated answers will be truthful or not.\footnote{This is a simpler, less computationally expensive version of the approach of~\citet{Kadavath2022-pf}, where multiple answers are sampled and the probe initially predicts a continuous score---proportion of correct answers in the sampled set.}

\noindent\textbf{Sequence Probability:} Average log-probability of the question's tokens.\footnote{This implementation captures how surprised the model is by the question, which is linked with expected correctness.}

\noindent\textbf{Verbalization:} Prompting for the confidence level of knowing the answer to the question (see Appendix~\ref{app:implementation} for details). 

\noindent\textbf{Surrogate Token Probability:} Log-probability of ``Yes'' token following a query on knowing the answer to the question.

\noindent\textbf{Consistency:} We prompt the model with the question and sample 10 responses with $\tau=1$. Then, we compute a matrix of pairwise NLI scores~\cite{Laurer2024-less} on all generations, and return an average.  

\subsection{Evaluating Scoring Methods}

To evaluate the methods, we use the area under the precision-recall curve (AUPRC), as is common in related literature, e.g.,~\citealt{Kadavath2022-pf}. %
Using a metric that considers various decision thresholds enables a robust comparison across methods. The higher AUPRC, the better ranking capability of the method, with cleaner separation between true/false statements or known/unknown facts. In an effort to make interpretation of AUPRC more intuitive for practical applications, we also report the precision and recall at K (Tables~\ref{tab:pr_pt}
and~\ref{tab:pr_pik} in Appendix~\ref{app:precision}).

\subsection{Models}
We study eight publicly available LLMs, with open access to  weights. 

We consider models with different sizes (7B to 46.7B), architecture, and training paradigms (instruction-fine-tuned or not) from the Falcon~\cite{almazrouei2023falcon} and Mistral~\cite{jiang2023mistral, jiang2024mixtral} model families (see Table \ref{tab:models}).

\begin{table}[t!]
\centering
\resizebox{\columnwidth}{!}{%
\setlength{\tabcolsep}{3pt}
\begin{tabular}{lllll}
\toprule
\textbf{Names} & \textbf{Size} & \textbf{Open} & \textbf{Arch.} & \textbf{Instruct} \\ \midrule
\textbf{Falcon} & 40B & \faCheck & Dense &  \\
\textbf{Falcon Inst.} & 40B & \faCheck & Dense & \faCheck \\
\textbf{Falcon} & 7B & \faCheck & Dense &  \\
\textbf{Falcon Inst.} & 7B & \faCheck & Dense & \faCheck \\
\textbf{Mixtral} & 46.7B & \faCheck & SMoE &  \\
\textbf{Mixtral Inst.} & 46.7B & \faCheck & SMoE & \faCheck \\
\textbf{Mistral} & 7B & \faCheck & Dense &  \\
\textbf{Mistral Inst.} & 7B & \faCheck & Dense & \faCheck \\
\bottomrule
\end{tabular}
}
\caption{The models used in our experiments. \textit{Dense} represents the usual transformer decoder architecture, while SMoE stands for Sparse Mixture of Experts~\cite{shazeer2017outrageously}, here made of 8 experts of ~7B. \textit{Instruct}. models have been instruction fine-tuned. Open models have publicly available weights.}
\label{tab:models}

\end{table}
\subsection{Paraphrasing and Translation} \label{subsec:para_and_trans_data}

To test methods robustness and to disentangle confidence over a fact from confidence based on a specific wording, we generate semantically equivalent variants of statements/questions from Lama T-REx and PopQA (see Section~\ref{sec:robustness}).
For each input, we generate 10 paraphrases by prompting Mixtral-8x7B-Instruct-v0.1 (prompt and examples are in Appendix~\ref{app:paraphrasing}).
We remove repetitions and only keep paraphrases that are semantically equivalent\footnote{To detect semantic equivalence, we test for entailment in both directions using an NLI model~\cite{Laurer2024-less}} to the original input.
This results in an average of eight paraphrases per original input.

We also consider translation as another meaning-preserving transformation. 
Specifically, we translate the English examples to two languages from different language families ({Romance} and Slavic):\footnote{For translations we use Amazon Translate.} (\emph{i})~French---a high resource language to which all models except the 7B Mistral models have been explicitly exposed to during training,
and (\emph{ii})~Polish---a language on which we expect lower degree of competence (Falcon reports ``limited capability'' for Polish~\cite{almazrouei2023falcon}, while Mistral models do not mention it at all).
Finally, we manually verified the quality of a sample of 100 translations, finding them to be meaning preserving and vastly without errors.

\section{Empirical Comparison of the Methods}
\label{sec:empirical-comp}
\subsection{\texorpdfstring{\pIK}~ on Lama T-REx} \label{subsec:pt}

\begin{figure}[t]
    \centering
    \hspace{-3.5mm}\includegraphics[width=1.04\columnwidth]{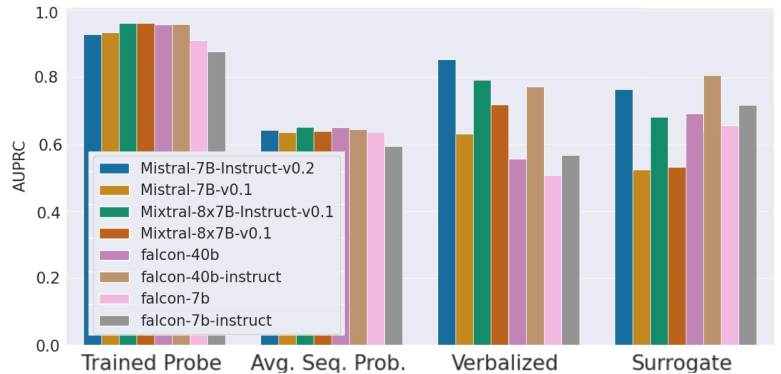}
    \caption{AUPRC scores on T-REx with both true and false statements;  \texorpdfstring{\pTrue}~.}
    \label{fig:p_t}
\end{figure}

With each of the four methods, we derive estimates of factual confidence for all statements in the Lama T-REx test set, repeating the experiment for each LLM. We evaluate the reliability of a method by checking whether it yields \pTrue scores that can effectively separate the true statements from the false statements, measured as AUPRC.

We report the results of this analysis in Figure~\ref{fig:p_t}. The \emph{trained probe} method performs best, outperforming the \emph{sequence probability} by an average AUPRC of .3. Of all methods and models, only the \emph{verbalized method} is  competitive {to the supervised probe}, and only for Mistral 7B instruct. Otherwise all methods perform at least .1 AUPRC below the \emph{trained probe}. 
This result suggests that information about the expected truth value of a statement is better captured in deeper layers of the network, as opposed to the output scores.

While for \emph{trained probe} and \emph{average sequence probability} we note relatively small differences in AUPRC across models, for the \emph{verbalized} and \emph{surrogate} methods we see large variation.
Concretely, instruction-tuned models always perform better than their counterparts. 
This is expected as both methods require to follow instructions in the prompt.

Model size also seems to have an effect: all 40B+ models perform better than their 7B counter-parts, with the exception of Mistral-7B-Instruct-v0.2. 

{Information on the specific differences between versions of Mistral is not publicly available, making the interpretation of this result difficult. One possibility is that the second version of the model has better instruction-following capabilities, but we cannot be sure (\citealt{behnamghader2024llm2vec} discusses conjectures on unusual properties of the Mistral family of models).}

Finally, while the \emph{average sequence probability} method performs consistently above chance (50\%), it has overall poor performance  in comparison to other methods. It only outperforms the other non-trained methods---\emph{verbalized} and \emph{surrogate}---for non-instruction-tuned models.

\subsection{\pIK on PopQA} \label{results:pik}
\begin{figure}[t]
    \centering
    \hspace{-3mm}\includegraphics[width=1.04\columnwidth]{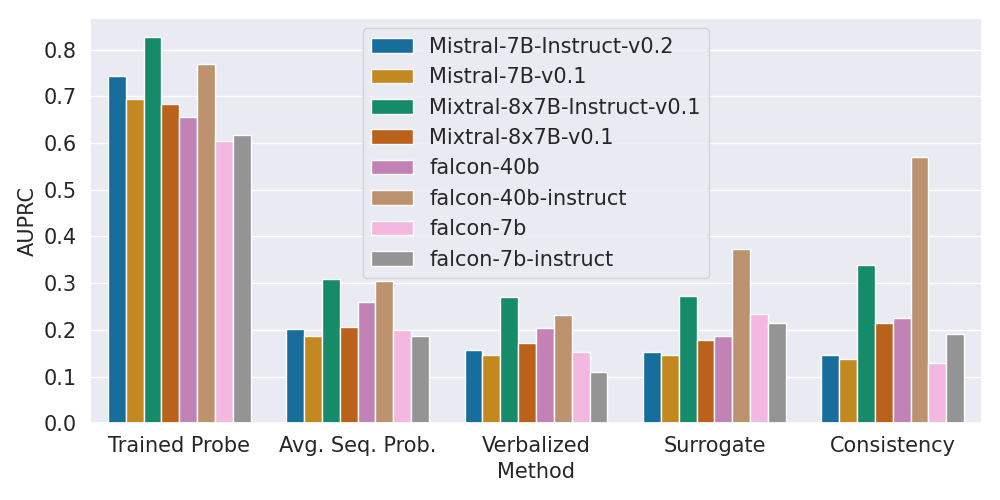}
    \caption{AUPRC scores on PopQA dataset; \texorpdfstring{\pIK}~.} 
    \label{fig:pik}
\end{figure}

\pIK estimates the degree of a model's confidence that its predicted answer will be correct.
A good estimator of \pIK would thus assign high scores to queries which the model answers correctly, and low scores to others. Following this reasoning, for \pIK 
we compute the AUPRC scores using binary labels that encode whether the model's answer (in our case, generated with greedy decoding) is correct.
Note that this way of computing AUPRC---based on \emph{future correctness}---provides a direct  estimate of the method's expected effectiveness for hallucination mitigation (when the method is used to automatically detect when the model should abstain from answering). 
In this scenario, a method is effective only if its estimates are actually predictive of the correctness of the model's answers. 

The results are reported in Figure~\ref{fig:pik}. In this experiment we also study the \emph{Consistency} method, which was omitted from \pTrue results because, by design, it cannot be applied to an entire statement.
Overall, \pIK is harder to estimate than \pTrue, with lower AUPRC results: e.g., The best trained probe is 0.1 below in AUPRC for \pIK than it is for \pTrue. 
This may be due to the complexity of the setup---in QA the confidence is estimated only based on a query, in contrast to fact-verification. But it may also be that the binary \emph{future correctness} labels used for our AUPRC computation introduce some noise. E.g., the model may be genuinely uncertain and still output the correct answer by chance.

The \emph{trained probe} method is again, by large, the most reliable across all models. 
With the exception of Falcon-40B instruct, the other methods perform close to or below chance (depending on the model's label distribution, chance level varies between 0.11 and 0.27).
This indicates that non-trained estimators are generally not reliable for \pIK despite being frequently used in the literature.
Within each method, we observe differences across models---up to a 40\% margin. This can be linked to (\emph{i})~whether a model is instruction-tuned (as noted for \pTrue) and (\emph{ii})~the model family---with more reliable scores for Mistral models than for Falcon models.

\subsection{Generalization of the Trained Probe}
\label{subsubsec:generalization}
The results above highlight the trained probe as the most reliable estimator for factual confidence---both for \pTrue and \pIK. However, in those experiments we trained and evaluated the models within the same domain, which leaves open questions about the probe's generalization capabilities.
We address this gap by evaluating the model from \ref{subsec:pt}, trained to estimate \pTrue from Lama T-REx data, on the PopQA dataset converted to test for \pTrue.
Specifically, we re-work the PopQA data for the fact-verification setup by turning question-answer pairs into (evenly distributed) true and false statements, using the template: ``\emph{The answer to [QUESTION] is [ANSWER]}''.\footnote{For true statements we use the gold answers from PopQA dataset. For false statements, we sample alternative answers from the same question class in the dataset; e.g.,~\textit{The answer to ``\emph{In which country is Washington?}'' is ``\emph{United States of America}''} vs.~``\emph{South Korea}''.}
We derive estimates for \pTrue on such statements using the probes trained on Lama T-REx, and compute AUPRC (Table \ref{table:generalization}).

\begin{table}[t]
\centering
\begin{tabular}{llcc}
\toprule
\textbf{Name} & \textbf{Size} & \textbf{AUPRC} & \textbf{$\Delta$}  \\ \midrule
\textbf{Falcon} & 40B & .80 & -.16\\
\textbf{Falcon Ins.} & 40B & .81 & -.15 \\
\textbf{Falcon} & 7B & .66 & -.25  \\
\textbf{Falcon Ins} & 7B & .59 & -.28 \\
\textbf{Mixtral} & 46.7B &.78 &  -.18\\
\textbf{Mistral} & 7B & .62 & -.31\\
\textbf{Mistral Ins} & 7B &.75& -.18\\
\bottomrule 
\end{tabular}
\caption{AUPRC on PopQA test set re-worked as true/false statements, using \pTrue estimates from probes trained on Lama T-REx. $\Delta$: difference of AUPRC with respect to that for Lama T-REx data (in-domain).}
\label{table:generalization}
\end{table}

Going from in-domain to out-of-domain test data (Lama vs.~PopQA), we observe AUPRC differences of min -.15 and max -.31. However, the scores remain in a high range of [.62, .81] indicating substantial generalization.
The LLMs for which the probe retains the least and the most reliability are Mistral-7B and Falcon-40B-instruct, respectively. Interestingly, these are also the models getting the least and the most answers right on PopQA in the QA setup (14\% and 23\%).
This suggests that the transferability of the probe may be affected by how challenging the out-of-domain dataset is to the model.
In the next sections, we provide further evidence of probe generalization by looking at whether and to what extent the AUPRC is affected by input paraphrasing and translation.

\section{Robustness to Linguistic Variations}
\label{sec:robustness}
In this section, we apply meaning-preserving linguistic variations to each input in order to:
(\emph{i})~Assess the robustness of methods. The expectation is that if a method is robust, it should produce equally reliable estimates (equally high AUPRC scores) across different input formulations (Section~\ref{subsec:method_robustness});
(\emph{ii})~Investigate the stability of an LLM's encoding of facts. The expectation is that if a fact is well abstracted, the factual confidence should be invariant to semantics-preserving changes in the input (Section~\ref{subsec:robustness_of_encoding}).

We consider two types of input variation: paraphrases and translations.

\subsection{Robustness of Methods} \label{subsec:method_robustness}
We study method robustness in both \pTrue and \pIK, using the same setup as before; in particular, we do not retrain the \textit{trained probe} and do not adapt the prompts in any way.\footnote{Note this also applies to translations; i.e.,~the trained probe is trained on English data only and we use English prompts to query the model about French/Polish inputs.} To test robustness on paraphrases, we generate 10 different paraphrase sets---each holding different formulations of the original inputs---and compute AUPRC on each set. We observe that AUPRC remains stable for all methods (absolute variation between 5\% and 10\%), indicating they are robust to paraphrasing. The most affected method is the \textit{trained probe} in the \pIK setting, but even here we only note up to a standard deviation of 3 percentage points (for Mistral-7B). Full result tables are in Appendix~\ref{app:robustness}.

For translations, we compute a separate AUPRC on the French and Polish versions of T-REx.
We find varying degrees of method transferability to new languages (above .5 for \textit{verbalized}, up to .91 for \textit{trained probe}). 
All methods generalize to both French and Polish above chance, except for (\emph{i})~\emph{verbalized confidence} and (\emph{ii})~\emph{surrogate logits} when applied to Mistral models (see Figure~\ref{fig:auprc_translations} in the Appendix).
Notably, the probes trained on English data remain to a large extent reliable (AUPRC for French: .73-.91; for Polish: .61-.91) on unseen languages---with 40B+ models and the instruction-tuned Mistral demonstrating the highest transferability.
This provides additional evidence for out-of-domain generalization of trained probes (Section~\ref{subsubsec:generalization}).
In particular, the probes can extract scores that are discriminative of true and false statements also from hidden states computed from inputs in a different language than the one used at training.
This suggests that the LLMs encode factual confidence in a similar way across languages, {which is remarkable, given the differences in exposure to training data from the three languages.}

\subsection{Robustness of Facts Encoding in LLMs} \label{subsec:robustness_of_encoding}
We hypothesize that, to robustly learn facts and minimize hallucinations, a model has to build stable abstractions over different types of relevant evidence from the training data. 
We also expect that if the model has built such a robust representation of a fact, this would lead to equal confidence under equivalent formulations of that fact. Inconsistent confidence would in turn indicate excessive reliance on surface-level features. %

\begin{figure}[!t]
    \centering
    \includegraphics[width=\columnwidth]{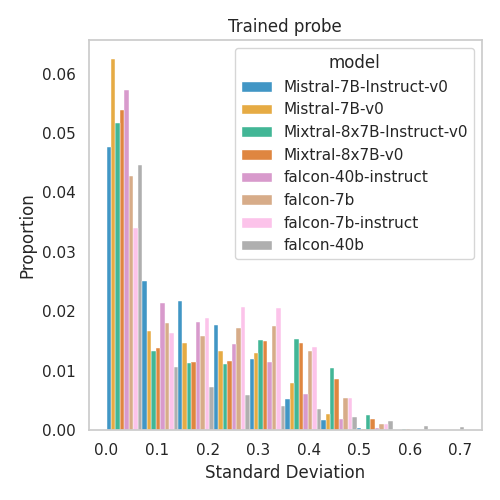}
    \caption{Distribution of standard deviation scores for normalized {\pTrue} on paraphrases of the same fact.
    }
    \label{fig:std_pt}
\end{figure}

Figure~\ref{fig:std_pt} shows how paraphrasing the input ($\sim$8 paraphrases per input causes changes in the \emph{trained probe}~\pTrue estimates across the T-REx dataset.\footnote{{We focus on the \emph{trained probe} since it was the best performing method.}} 
Specifically, we present the distribution of the standard deviation of confidence scores, across different facts.
The amount of variation is not stable across facts. While for each model, we observe a large amount of facts for which there is little-to-no variation in the score (indicating a stable  encoding), we also note many facts for which different wordings lead to strongly varying degrees of confidence (up to .5 standard deviation, with very few cases at .7).
This indicates inconsistent LLM behavior, with excessive sensitivity to how a claim is worded.
{Out of all models, Falcon-7B-instruct appears to have developed the least stable encoding, with the standard deviation distribution shifted towards higher values. On the other extreme, Mistral-7b-v0 appears to be the model showing less variation.}

To test robustness of factual knowledge across languages, we compare the distributions of \pTrue scores over the same facts using the Spearman correlation analysis (for language pairs) and the Friedman test (for language groups). %
Analysis reveals high correlations (Spearman's $\rho$ > .7; full results in Table~\ref{table:corr_translations} in Appendix) between factual confidence scores on all language pairs for the 40B+ models. 
In particular, we note the highest correlations (in the .87-.92 range) for Falcon 40B models, which points to highly robust multilingual behavior. However, the Friedman tests reveal that for all models, the differences across the distributions are statistically significant {(p-values close to 0)}; i.e., the differences in scores across the languages are not close enough to be coming from the same population. Given those results, we conclude that while there is a link between the confidence scores across the languages, this is not fully systematic.

\section{Discussion \& Conclusion}
\label{sec:ccl}
In this paper, we compare existing methods to estimate LLMs factual confidence. 
Obtaining reliable estimates %
can benefit LLMs applications, by anticipating potential hallucinations and limiting the non-factual information output by a model~\citep{tonmoy2024comprehensive,evans2021truthful}. %
However, if not reliable, such estimates can be counterproductive, as they would introduce errors and  negatively affect user-model interactions.

Our experiments across eight LLMs demonstrate that the \textit{trained probe} method is the most reliable estimator of LLM factual confidence. %
It works well for both fact-verification (\pTrue) and Question Answering (\pIK) across all 8 tested models, indicating that its reliability is likely to generalize to other LLMs.
{Moreover, we show that it generalizes to out-of-domain data: (\emph{i})~when a model trained on T-REx is applied to an unseen dataset (PopQA, Section~\ref{subsubsec:generalization}), and (\emph{ii})~in a cross-lingual transfer setting (Section~\ref{subsec:method_robustness}).}

{We must note, though, that the fine-tuning nature of the method clearly puts it at an advantage over the other zero-shot probing methods.}
Moreover, applying the \emph{trained probe} method comes with strong requirements: (\emph{i})~access to model weights---not always provided by proprietary LLMs, and (\emph{ii})~need of supervised data.
If these requirements cannot be met, but the model is instruction-tuned~\citep{ouyang2022training}
we recommend estimating \pTrue with \emph{verbalized confidence} or \textit{surrogate probabilities}. 

The other methods under study, especially if applied to non-instruction-tuned LLMs, are not consistently reliable. 

Our results highlight the need for more research to develop reliable estimators that can be applied to black-box models, with inaccessible internal representations. 
We expect the reliability gap between methods like \emph{verbalized confidence} and \textit{trained probe} to get smaller with increasingly powerful LLMs, especially in their ability to follow instructions.
However, strong results of \textit{trained probe} indicate that hidden states contain signal about factual confidence, and it is unclear whether this can ever be fully leveraged by the prompting approaches.

Besides the comparison among methods, we also provide insights on the stability of factual knowledge in LLMs~\citep{Petroni2019-pa,doi:10.1073/pnas.2215907120,mahowald2024dissociating}. 
We show that the factual confidence of an LLM is not always consistent under meaning-preserving variations of the input (paraphrases and translations). While the model may sometimes 
be sure that a fact is true or false, or that it knows the answer to a question, it may actually behave differently if we reformulate the statement or question.
An interesting direction for future research is the exploration of training methods that teach an LLM to better disentangle facts from the diversity of forms they can be stated in, and ultimately exhibit better and more consistent factual knowledge. 
This would also contribute to increasing LLMs resistance to adversarial attacks~\citep{madry2018towards}, mitigating the generation of misinformation due to an incorrect sensitivity to input changes.

\section*{Acknowledgments}
We thank the anonymous reviewers for their helpful questions and comments, which have helped us improve the quality of the paper. We also want to thank Ionut Sorodoc, Neha Anna John, Phu Mon Htut and the entirety of the AWS Bedrock Responsible AI team for the time they took to support, discuss and improve this work. We thank Marco Baroni and Nathana\"{e}l Rakotonirina for helpful discussion during this project.

Mat\'{e}o Mahaut
is supported by the European Research Council (ERC) as part of a project in the European Union’s Horizon 2020 research and innovation program (grant agreement No. 101019291). Views and opinions expressed are however those of the authors only and do not necessarily reflect those of the European Union or the European Research Council Executive Agency. Neither the European Union nor the granting authority can be held responsible for them.

\section*{Limitations}
Given the extensive scope of this work (eight models, five methods and two facets of factual knowledge), we did not have the capacity to study more complex aspects of factual confidence, such as non-atomic facts, reasoning or in-context learning. While our results show that the \emph{trained probe} is much stronger than other methods on T-REx and PopQA, there is no guarantee that this remains the case in more complex settings. Furthermore, methods themselves have limitations, making comparison use-case dependent. The \emph{trained probe} method for example requires training data, and while we have tested for transfer capabilities in our simple atomic fact setup,~\citet{Kadavath2022-pf} have shown that there are limits to the kind of tasks this method can be transferred to. The same can be said of the \textit{sequence probability} method, which in our experiments works better than both prompt-based methods for non instruction fine-tuned models. While this method performs well on simple atomic facts, more complex sentences, or even simple but longer sentences could lead to weaker results. Furthermore, both prompt-based methods are sensitive to prompt-variations. %

\section*{Ethics and Broader Impact}
This work contributes to the wider goal of automatically reducing risk when using LLMs. We contribute to false statement detection and answer confidence, leading to potential applications which can build trust in LLMs. None of the methods studied completely solve the issue of hallucination, or non-factual utterances of models, leaving a need for future works on the subject. While methods studied can work with models with 7B and 40B+ parameters, the deployment of those models requires specific infrastructure and is compute intensive.

\bibliography{custom}

\clearpage
\appendix
\section*{Appendix for ``Factual Confidence of LLMs: on Reliability and Robustness of Current Estimators''}

\section{Implementation of scoring methods}
\label{app:implementation}
\subsection{Verbalized}
For the \emph{verbalization} method, we start with the prompt from~\citet{Tian2023-wp} and adapt it to the current tasks. Then, we filter out non-numerical outputs and normalize.

\noindent Prompt used for \pTrue experiments: 
\begin{quote}\textit{Provide your confidence level (on a scale of 1.0 to 10.0) that the following statement is correct}  .\\\\
\textit{The statement is: \$statement} \\\\
\textit{Confidence level:}
\end{quote}

\noindent Prompt used for \pIK experiments: 
\begin{quote}\textit{Provide your confidence level (on a scale of 1.0 to 10.0) that you can answer the following question} .\\\\
\textit{The question is: \$question} \\\\
\textit{Confidence level:}\end{quote}

\subsection{Surrogate token probabilities}
For \pTrue we use the prompt: \begin{quote}\textit{Q: Is the following statement true: \$statement (Yes/No/Maybe)?} \\
\textit{A: } \end{quote}
For \pIK we use the prompt: \begin{quote} \textit{Q: Do you know the answer to the following question: \$question (Yes/No/Maybe)?} \\
\textit{A: }\end{quote}
\subsubsection{Consistency}

The ground truth answer is generally a word, ranging between two and five tokens. We output 10 samples of up to 25 tokens using a temperature ($\tau$) of 1, leaving ample room for the ground truth to be generated in a sentence. Finally, we calculate the consistency score by computing the mean NLI~\cite{Laurer2024-less} score of these 10 samples.

\section{Paraphrasing}
\label{app:paraphrasing}

Prompt used to generate paraphrases with Mixtral-8x7B-Instruct-v0.1 (examples are provided in Table~\ref{tab:ex_paraphrases}): 

\begin{quote}
    \textit{Given a sentence, generate paraphrases of it as follows:}
    \begin{quote}\textit{- You can change and/or add words, and/or change the syntactic structure of the sentence;}
    \end{quote}
    \begin{quote}
    \textit{- Make sure the new sentence does not add additional details with respect to the original.}
    \end{quote}
    \begin{quote}\textit{- Make sure the new sentence does not omit any details with respect to the original.}
    \end{quote}
    \begin{quote}
    \textit{- Make sure the new sentence is natural and plausible, in spite of the changes.}
    \end{quote}
    \begin{quote}
    \textit{- Do not generate the original sentence or previously generated ones.}
    \end{quote}
    \textit{List your paraphrases as bulletpoint.}\\
    \textit{Sentence: \$sentence}\\
    \textit{New sentences:}
\end{quote}

\begin{figure*}[!t]
    \centering
\includegraphics[width=0.9\textwidth]{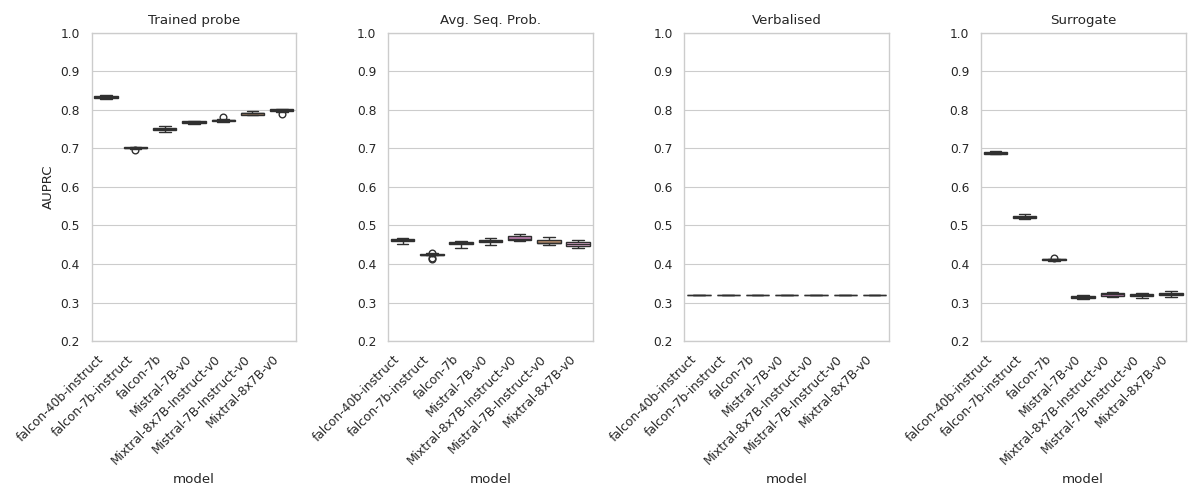}
\caption{Variation in \pTrue AUPRC when sampling paraphrases. 10 sets of paraphrases are randomly sampled, with one paraphrase for every question in Lama Lama T-RE.}
    \label{fig:auprc_stab_pt}
\end{figure*}

\begin{figure*}[!t]
    \centering
    \includegraphics[width=0.9\textwidth]{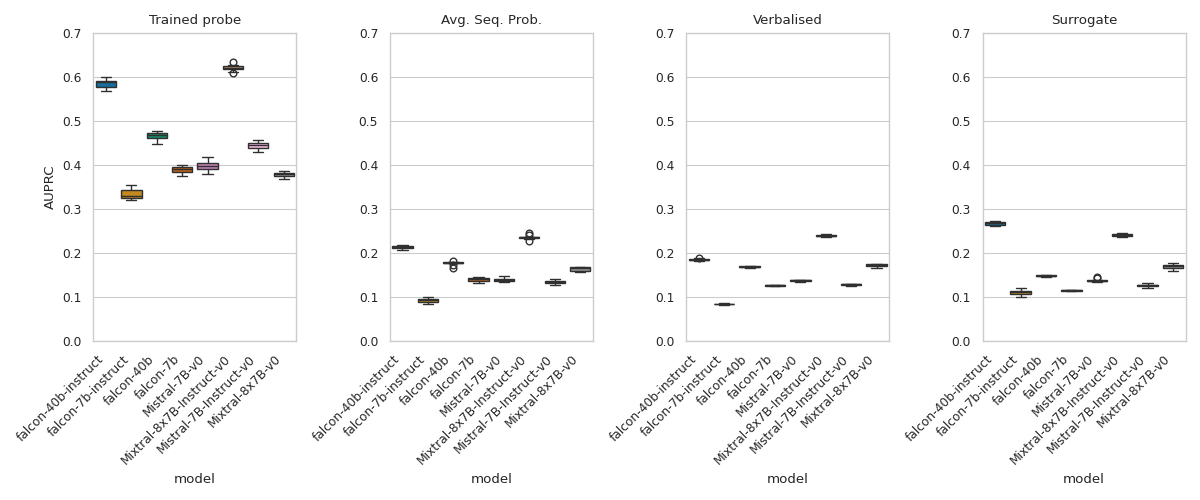}
    \caption{Variation in \pIK AUPRC when sampling paraphrases. 10 sets of paraphrases are randomly sampled, with one paraphrase for every question in PopQA.}
    \label{fig:auprc_stab}
\end{figure*}

The variation of \pTrue and \pIK with paraphrases are provided in Figures~\ref{fig:auprc_stab_pt} and~\ref{fig:auprc_stab}.

\begin{table}[t!]
\centering
\begin{tabular}{llcc}
\toprule
\textbf{Name} & \textbf{Size} & \textbf{En-Fr} & \textbf{En-Po}  \\ \midrule
\textbf{Falcon} & 40B  & .90 & .86\\
\textbf{Falcon Ins.} & 40B & .92 & .87 \\
\textbf{Falcon} & 7B &  .79 & .44 \\
\textbf{Falcon Ins} & 7B & .67 & .35 \\
\textbf{Mistral} & 7B & .67 & .58\\
\textbf{Mistral Ins} & 7B &.65& .53\\
\textbf{Mixtral} & 46.7B &.87 &  .77\\
\bottomrule 
\end{tabular}
\caption{Spearman correlation coefficient for English-French and English-Polish \pTrue scores on translated Lama T-REx statements. }
\label{table:corr_translations}
\end{table}

\section{Method Robustness to Variation}
\label{app:robustness}
To measure the robustness of the methods towards linguistic variations, we randomly sample a paraphrase for every sentence in the original dataset, making ten sets of paraphrases of the same size. We then compute AUPRC without changing the method in any way for the ten sets, and calculate the variance in results. The results are shown in Figures~\ref{fig:auprc_stab_pt} and~\ref{fig:auprc_stab}.
All methods remain stable, and robust to paraphrases. The biggest variation occurs for the \textit{trained probe method}, but are only of the order of 3 percentage points. 

\begin{figure*}[!t]
    \centering
\includegraphics[width=0.9\textwidth]{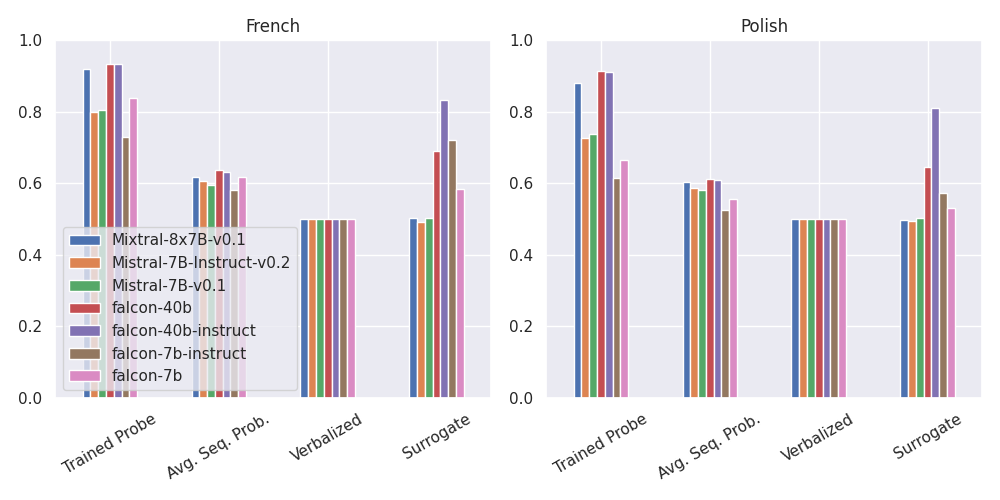}
    \caption{AUPRC for \pTrue scores for translations in French and Polish of the Lama T-REx statements. }
    \label{fig:auprc_translations}
\end{figure*}

Table~\ref{table:corr_translations} shows the correlation between scores across different languages, and Figure~\ref{fig:auprc_translations} shows AUPRC of all four methods evaluated on the French and Polish versions of the Lama T-REx dataset.

\begin{table*}[t]
\centering
\begin{tabular}{llllllllllllll}
\toprule
\textbf{Name} & \textbf{Size} & \multicolumn{3}{l}{\textbf{Surrogate}} & \multicolumn{3}{l}{\textbf{Trained Probe}} & \multicolumn{3}{l}{\textbf{Avg. Seq. Prob}} & \multicolumn{3}{l}{\textbf{Verbalized}} \\ \cline{3-14} 
 &  & \textit{r90} & \textit{r70} & \textit{r50} & \textit{r90} & \textit{r70} & \textit{r50} & \textit{r90} & \textit{r70} & \textit{r50} & \textit{r90} & \textit{r70} & \textit{r50} \\
 \midrule
\textbf{Falcon} & \multicolumn{1}{l|}{40B} & .33 & .34 & \multicolumn{1}{l|}{.45} & .60 & .77 & \multicolumn{1}{l|}{.87} & .36 & .40 & \multicolumn{1}{l|}{.45} & .32 & .32 & .32 \\
\textbf{Falcon Ins.} & \multicolumn{1}{l|}{40B} & .43 & .57 & \multicolumn{1}{l|}{.69} & .62 & .78 & \multicolumn{1}{l|}{.88} & .36 & .41 & \multicolumn{1}{l|}{.46} & .32 & .32 & .32 \\
\textbf{Falcon} & \multicolumn{1}{l|}{7B} & .32 & .32 & \multicolumn{1}{l|}{.33} & .49 & .64 & \multicolumn{1}{l|}{.77} & .36 & .40 & \multicolumn{1}{l|}{.44} & .32 & .32 & .32 \\
\textbf{Falcon Ins.} & \multicolumn{1}{l|}{7B} & .34 & .37 & \multicolumn{1}{l|}{.41} & .60 & .76 & \multicolumn{1}{l|}{.87} & .34 & .37 & \multicolumn{1}{l|}{.41} & .32 & .32 & .32 \\
\textbf{Mixtral} & \multicolumn{1}{l|}{46.7B} & .32 & .32 & \multicolumn{1}{l|}{.31} & .59 & .75 & \multicolumn{1}{l|}{.84} & .35 & .39 & \multicolumn{1}{l|}{.44} & .32 & .32 & .32 \\
\textbf{Mixtral Ins.} & \multicolumn{1}{l|}{46.7B} & .32 & .31 & \multicolumn{1}{l|}{.30} & .60 & .74 & \multicolumn{1}{l|}{.81} & .36 & .41 & \multicolumn{1}{l|}{.46} & .32 & .32 & .32 \\
\textbf{Mistral} & \multicolumn{1}{l|}{7B} & .32 & .32 & \multicolumn{1}{l|}{.32} & .53 & .67 & \multicolumn{1}{l|}{.80} & .35 & .40 & \multicolumn{1}{l|}{.44} & .32 & .32 & .32 \\
\textbf{Mistral Ins.} & \multicolumn{1}{l|}{7B} & .32 & .32 & \multicolumn{1}{l|}{.32} & .56 & .71 & \multicolumn{1}{l|}{.83} & .36 & .41 & \multicolumn{1}{l|}{.45} & .32 & .31 & .30 \\
\bottomrule
\end{tabular}
\caption{Method precision for estimating \pTrue on the Lama T-REx dataset for 3 recall values (P@90, P@70, P@50). Here, we set a threshold to ensure a certain recall, and look at the resulting precision. A recall of 90 with .72 precision would mean that when we select a score threshold that ensures 90\% of True sentences are correctly classified as such, 72\% of all sentences in the tested dataset are correctly classified. }
\label{tab:pr_pt}
\end{table*}

\begin{table*}[t]
\centering
\begin{tabular}{llllllllllllll}
 \toprule
\textbf{Name} & \textbf{Size} & \multicolumn{3}{l}{\textbf{Surrogate}} & \multicolumn{3}{l}{\textbf{Trained Probe}} & \multicolumn{3}{l}{\textbf{Avg. Seq. Prob.}} & \multicolumn{3}{l}{\textbf{Verbalized}} \\ \cline{3-14} 
 &  & \textit{r90} & \textit{r70} & \textit{r50} & \textit{r90} & \textit{r70} & \textit{r50} & \textit{r90} & \textit{r70} & \textit{r50} & \textit{r90} & \textit{r70} & \textit{r50} \\
\midrule
\textbf{Falcon} & \multicolumn{1}{l|}{40B} & .33 & .34 & \multicolumn{1}{l|}{.45} & .60 & .77 & \multicolumn{1}{l|}{.87} & .36 & .40 & \multicolumn{1}{l|}{.45} & .32 & .32 & .32 \\
\textbf{Falcon Ins.} & \multicolumn{1}{l|}{40B} & .26 & .36 & \multicolumn{1}{l|}{.40} & .27 & .37 & \multicolumn{1}{l|}{.48} & .17 & .18 & \multicolumn{1}{l|}{.18} & .17 & .17 & .17 \\
\textbf{Falcon} & \multicolumn{1}{l|}{7B} & .11 & .11 & \multicolumn{1}{l|}{.11} & .20 & .26 & \multicolumn{1}{l|}{.31} & .11 & .11 & \multicolumn{1}{l|}{.12} & .11 & .11 & .11 \\
\textbf{Falcon Ins.} & \multicolumn{1}{l|}{7B} & .07 & .07 & \multicolumn{1}{l|}{.07} & .20 & .26 & \multicolumn{1}{l|}{.27} & .07 & .08 & \multicolumn{1}{l|}{.08} & .07 & .07 & .07 \\
\textbf{Mixtral} & \multicolumn{1}{l|}{46.7B} & .16 & .16 & \multicolumn{1}{l|}{.16} & .18 & .24 & \multicolumn{1}{l|}{.29} & .16 & .15 & \multicolumn{1}{l|}{.15} & .16 & .16 & .16 \\
\textbf{Mixtral Ins.} & \multicolumn{1}{l|}{46.7B} & .21 & .21 & \multicolumn{1}{l|}{.22} & .23 & .28 & \multicolumn{1}{l|}{.37} & .21 & .21 & \multicolumn{1}{l|}{.21} & .21 & .21 & .21 \\
\textbf{Mistral} & \multicolumn{1}{l|}{7B} & .12 & .12 & \multicolumn{1}{l|}{.12} & .14 & .21 & \multicolumn{1}{l|}{.25} & .12 & .12 & \multicolumn{1}{l|}{.12} & .12 & .12 & .12 \\
\textbf{Mistral Ins.} & \multicolumn{1}{l|}{7B} & .11 & .12 & \multicolumn{1}{l|}{.12} & .15 & .19 & \multicolumn{1}{l|}{.27} & .11 & .12 & \multicolumn{1}{l|}{.12} & .12 & .12 & .12 \\
\bottomrule
\end{tabular}
\caption{Method precision for estimating \pIK on the PopQA dataset at for 3 recall values (P@90, P@70, P@50)}
\label{tab:pr_pik}
\end{table*}

\section{Analysis of specific Precision and Recall}
\label{app:precision}
In Tables~\ref{tab:pr_pt} and~\ref{tab:pr_pik} respectively, we show results for  precision at different recall thresholds for the two formations: \pTrue on Lama T-REx and \pIK on PopQA. Similar to our previous findings (see Figures~\ref{fig:p_t} and~\ref{fig:pik}) the \emph{trained probe} noticeably outperforms other methods, making the differences much more apparent, confirming that for \pIK the Falcon 40B outperforms all other models, and all methods, expected trained probe, show low performance when applied with Falcon 7b Instruct. 
In constrast to the other methods, the precision of the \emph{verbalized confidence} does not change for different recall thresholds, for both \pTrue and \pIK, which suggests that we are reaching its limits to estimate factual confidence.
We hypothesize that the emph{verbalized confidence} failing to disentangle correctly \emph{True} sentences from \emph{False} ones.
This is also true, to some extent, for the \emph{surrogate token} method and the \emph{sequence probability} for \pIK. The \emph{trained probe}, on the other hand, has better precision with lower recalls, which is also the case for the \emph{sequence probability} for \pIK, as well as the \emph{surrogate token} method in the same condition, however only with bigger Falcon models. These results do push for additional work, as they point out that there remains substantial overlap when classifying True and False sentences (as well as successfully or unsuccessfully completed sentences) with a maximum of around 60\% of sentences correctly classified, reaching a recall of 90 in both settings.

\begin{table*}
\centering
\begin{tabular}{l|l}
\toprule
\textbf{Original sentence} & \textbf{Paraphrases} \\ \midrule
\multicolumn{2}{c}{\textbf{Lama T-REx}}  \\
\midrule
\multirow{5}{*}{Michie Mee is a actress by profession.} & Acting is the profession of Michie Mee. \\
 & Michie Mee makes a living as an actress. \\
 & Michie Mee is a professional actress. \\
 & Michie Mee is an actress in her profession. \\
 & Michie Mee is an artist who acts for a living. \\ \midrule
\multirow{6}{*}{\begin{tabular}[c]{@{}l@{}}The Munsters was originally aired on\\ Bravo network .\end{tabular}} & Bravo network was the first to air The Munsters. \\
 & The Munsters was first shown on Bravo. \\
 & The Munsters was first transmitted on Bravo. \\
 & Bravo was the first network to air The Munsters. \\
 & The Munsters was first broadcasted on Bravo. \\ \midrule
\multicolumn{2}{c}{\textbf{PopQA}}  \\ \midrule
\multirow{7}{*}{What is George Rankin’s occupation?} & What does George Rankin do for a living? \\
 & What line of work is George Rankin in? \\
 & What is George Rankin’s job? \\
 & What is George Rankin’s profession? \\
 & Can you tell me what George Rankin does? \\
 & George Rankin’s employment, could you tell me about it? \\
 & George Rankin’s work, what is it? \\ \midrule
\multirow{5}{*}{In what city was Louis Renault born?} & Where did Louis Renault come into the world? \\
 & In which urban area did Louis Renault enter the world? \\
 & In what metropolis did Louis Renault make his appearance? \\
 & In which city did Louis Renault first see the  light of day? \\
 & In which city was Louis Renault given birth? \\ \bottomrule
\end{tabular}
\caption{Examples of automatic paraphrasing from the T-REx and PopQA datasets.}
\label{tab:ex_paraphrases}
\end{table*}
\end{document}